\documentclass[sn-mathphys]{sn-jnl}

\jyear{2021}%

\theoremstyle{thmstyleone}%
\theoremstyle{thmstyletwo}%

\theoremstyle{thmstylethree}%

\usepackage{textcomp}
\usepackage{amssymb}
\usepackage{pifont,calc}
\usepackage{graphicx}
\usepackage[caption=false]{subfig}
\usepackage{caption,setspace}

\usepackage{amsfonts}
\usepackage{array}

\usepackage{cases} 
\usepackage{bm} 
\usepackage{mathrsfs}
\usepackage{url}
\usepackage{algorithm}
\usepackage{algpseudocode}

\usepackage{arydshln}
\usepackage{multirow}
\usepackage{multicol}

\makeatletter

\newcommand{\Rmnum}[1]{\expandafter\@slowromancap\romannumeral #1@}
\makeatother

\begin{document}

\title[Article Title]{An Efficient Implementation to Compute the Pseudoinverse for the Incremental Broad Learning System
on Added Inputs}


\author[1]{\fnm{Hufei} \sur{Zhu}}\email{hufeizhu93@hotmail.com}

\author*[2]{\fnm{Zhulin} \sur{Liu}}\email{liuzhl@scut.edu.cn}

\author[2,3]{\fnm{C. L. Philip} \sur{Chen}}\email{philipchen@ieee.org}

\author[1]{\fnm{Yanyang} \sur{Liang}}\email{liangyanyang@163.com}


\affil*[1]{\orgdiv{Faculty of Intelligent Manufacturing}, \orgname{Wuyi University}, \orgaddress{\city{Jiangmen}, \postcode{529020}, \state{Guangdong}, \country{China}}}

\affil[2]{\orgdiv{School of Computer Science and Engineering}, \orgname{South China University of Technology}, \orgaddress{\city{Guangzhou}, \state{Guangdong}, \country{China}}}

\affil[3]{\orgdiv{Faculty of Science and Technology}, \orgname{University of Macau}, \orgaddress{\city{Macau},  \state{Macau}, \country{China}}}


\abstract{In this paper, we improve the broad learning system (BLS)
by speeding up the incremental learning  for added inputs.
We propose an efficient implementation  for a step
that is in the pseudoinverse computation of
 a  partitioned matrix, to  reduce the  computational complexity.
 The proposed efficient implementation has two different forms for the cases
 of $q > k$ and $q \le k$, respectively, where $q$  and $k$ denote  the  number of additional training samples
  and the total number of  nodes, respectively.
  The proposed
 implementation for
   $q > k$ utilizes   the inverse of a sum of matrices
 to
 compute only a $k \times k$ matrix inverse, instead of a $q \times q$ matrix inverse
 in the original implementation,
 and the corresponding speedup for the matrix inversion operation  in the number of floating-point operations is $\frac{1}{2}(q/k)^3$.
 Moreover,
it
      also
 speeds up two relevant matrix multiplication operations in
 the original   implementation.
 On the other hand,
   the proposed
 implementation for $q \le k$ speeds up one matrix multiplication operation in
 the original   implementation.
 The numerical simulations
  show that both the proposed and original implementations
  always  achieve the same testing accuracy. On the Modified National Institute of Standards and Technology
dataset, the speedups of the proposed efficient BLS implementation over the original BLS implementation in total training time
  are 1.35-1.57 when $q>k$ and  1.09-1.13 when $q<k$, while on the NYU object recognition
benchmark dataset, the speedups  are 1.10-1.35 when $q>k$ and  1.07-1.09 when $q<k$.}




\maketitle

\section{Introduction}\label{sec1}

Single
 layer feedforward neural networks (SLFN) with universal approximation capability have been
widely applied in classification and
regression~\cite{BL_Ref_18,BL_Ref_19,BL_Ref_20}.
Traditional  methods to train SLFN, i.e., Gradient-descent-based learning algorithms~\cite{BL_Ref_22,BL_Ref_23},
converge slowly. They may halt at a local minimum, and
their generalization performance is sensitive to the
training parameters (e.g., learning rate). As a different  method to train SLFN,
the random vector functional-link neural network (RVFLNN)~\cite{BL_Ref_19}
has fast learning speed, and offers the generalization capability in function approximation~\cite{BL_Ref_20}.
In RVFLNN, only the output weights are trained, while the input weights and the biases are
randomly generated.

For a new added node or input,
a dynamic
step-wise updating algorithm was proposed in \cite{27_ref_BL_trans_paper} to update
the output weights of the RVFLNN easily,
by only computing the
pseudoinverse of the corresponding added node or input.
Then in \cite{BL_trans_paper},  broad learning system (BLS) was proposed to improve the previous scheme~\cite{27_ref_BL_trans_paper} in three aspects.
Firstly, BLS generates the mapped features from the input data to form the feature nodes,
which are enhanced as the enhancement nodes with random weights, and finally it
feeds the connections of all the feature
and enhancement nodes into the output.
 Secondly, BLS updates
the output weights easily
 for any number of new added nodes or
inputs,
by utilizing only one iteration to compute the
pseudoinverse of the corresponding added nodes or inputs.
 Lastly,  the ridge regression approximation of the pseudoinverse is applied to compute
the output weights for BLS, to achieve a better generalization performance.

%
%
%
%

 In \cite{BL_trans_paperApproximate}, the universal approximation capability of BLS
has been proven
mathematically,
and several BLS variants were proposed, which include cascade, recurrent, and broad-deep combination structures.
Then
more BLS variants have been proposed for various specific scenarios~\cite{BLS_work_2,IEEEaccessRef1,IEEEaccessRef2,IEEEaccessRef6,IEEEaccessRef7,IEEEaccessRef12,IEEEaccessRef3}.
A novel recurrent BLS  was proposed in \cite{BLS_work_2}, which connects the enhancement nodes recurrently to capture
the dynamic characteristics of a time series.
In \cite{IEEEaccessRef1},
a hidden layer  has been added on the enhancement nodes of BLS,  to solve the problem of small samples by shallow learning of the enhanced features.
In \cite{IEEEaccessRef2},  a BLS scheme was proposed
for object detection using the event data,  which includes a gradient descent algorithm to train network parameters.
Since the defect of BLS includes the sensibility to the number of feature nodes,
 a probabilistic model called sparse Bayesian BLS (SBBLS) to dispose of this defect was proposed in \cite{IEEEaccessRef6},  to linearly-weight a small group of basis functions from an enormous number of candidates within Bayesian framework.
 Broadly fused feature representation was utilized in \cite{IEEEaccessRef7} to improve the ability of BLS,
where a cost-sensitive BLS framework was proposed for imbalanced learning from disproportionate size of categories instances,
 and greedy BLS (GBLS) was proposed in \cite{IEEEaccessRef12} to deal with the redundancy of the hidden layer, of which the  structure  can be regarded  as a combination of unsupervised multi-layer feature representation and supervised classification or regression.
The unified framework called Domain Transfer BLS (DTBLS) was proposed in \cite{IEEEaccessRef3},
to address the issue of drift via adaptive compensation.

BLS, as a typically dynamic neural networks, has been widely applied in computer vision~\cite{BLS_work_1,BLS_work_5,BLS_work_6,IEEEaccessRef14,IEEEaccessRef16,IEEEaccessRef5,IEEEaccessRef4,IEEEaccessRef13}.
The Takagi-Sugeno  fuzzy system was merged into BLS to propose the fuzzy BLS in  \cite{BLS_work_1}, which was applied to synthesize multiview HDR (high dynamic range) images in  \cite{BLS_work_5}.
In  \cite{BLS_work_6},  BLS has been
 utilized  to estimate sun visibility for outdoor shadow estimating.
 A novel baseline model training scheme with
BLS-based incremental learning was developed in \cite{IEEEaccessRef14}
 for sensor-based human activity recognition,
  and a Fourier BLS is applied in \cite{IEEEaccessRef16} to learn satellite telemetry data with less computational and time resources.
Based on BLS,
a broad graph-based robust continuous clustering algorithm was proposed in \cite{IEEEaccessRef5} to upgrade the robust continuous clustering (RCC),
and
 a new semi-supervised classification algorithm was developed in \cite{IEEEaccessRef4}
 to improve the clustering performance of RCC.
In \cite{IEEEaccessRef13}, BLS was applied with local feature fusion to propose a fast training method for facial beauty prediction.

 On the other hand, many fault diagnosis methods based on BLS have been developed~\cite{BLS_work_3,IEEEaccessRef17,IEEEaccessRef8,IEEEaccessRef9,IEEEaccessRef10,IEEEaccessRef15,IEEEaccessRef11}.
The weighted BLS proposed in \cite{BLS_work_3}
tackles the noise and outliers in an industrial process,
by assigning lower weights
to the  abnormal samples
for the purpose of decreasing their contributions,
while  the fault aware BLS  proposed in \cite{IEEEaccessRef17}
reduces the effect of weight/node failures on the BLS, which is based on an objective function for enhancing the fault aware performance.
BLS has also been  utilized in  \cite{IEEEaccessRef8}
to distinguish abnormities under large-scale pipeline network environments.
BLS was applied in \cite{IEEEaccessRef9} to develop
an efficient and responsive motor fault diagnostic method,
and  was utilized in \cite{IEEEaccessRef10}
to diagnose TPIM (three-phase induction motor) faults.
Moreover, an efficient online model based on BLS was  establish in \cite{IEEEaccessRef15} for detecting anomalies in the braking system,
and a new fault diagnosis (PABSFD) method based on BLS was proposed in \cite{IEEEaccessRef11} for rotor system,
 which utilizes the principal component analysis (PCA).






Despite the success of broad learning in dynamic construction, its applications would also suffer from efficiency shrinkage. The more efficient increamental algorithms are nessarry for above applications. This paper focuses on the incremental BLS on added inputs. We  reduce the computational complexity of the incremental learning  for added inputs
in \cite{BL_trans_paper},
by proposing an efficient implementation
for a step in the pseudoinverse computation of
 a  partitioned matrix.
 The proposed implementation has two different forms
for the cases of $q > k$ and $q \le k$, respectively, where $q$  and $k$ denote  the  number of additional training samples
  and the total number of  nodes, respectively.
  When $q > k$, the proposed  implementation utilizes the inverse of a sum of matrices~\cite{InverseSumofMatrix8312}
  to
 compute only a $k \times k$ matrix inverse instead of a $q \times q$ matrix inverse
 required in the original  BLS~\cite{BL_trans_paper},
 and  also
 speeds up two relevant matrix multiplication operations in
   the original  BLS.
 When $q \le k$, the proposed
 implementation speeds up one matrix multiplication operation in
 the original BLS.
Compared to the original BLS on added inputs~\cite{BL_trans_paper},
the proposed efficient implementation of BLS theoretically doesn't change the performance since it only reduces the computational complexity.


In Section~\ref{sec2},
the original incremental BLS on added inputs are introduced.
Then in  Section~\ref{sec3},
 we  propose an efficient implementation
for a step in the pseudoinverse computation of
 a  partitioned matrix,
 to speed up
the original  incremental BLS on added inputs.
 Section~\ref{sec4}
 compares the expected computational complexities of
 the
BLS with the proposed implementation and
the original BLS,
and evaluates
them
 by numerical experiments.
 Finally we conclude this paper in Section~\ref{sec5}.


\section{Original Incremental Broad   Learning System on Added  Inputs}\label{sec2}

RVFLNN enhances the input data ${{\mathbf{X}}}$ directly, to obtain the enhancement components
$\xi ({{\mathbf{X}}}{{\mathbf{W}}_{{{h}}}}+{{\bm{\beta }}_{{{h}}}})$
with the  weights ${{\mathbf{W}}_{{{h}}}}$
and biases ${{\bm{\beta }}_{{{h}}}}$ to be random.
The input data and the enhancement components form
the expanded input matrix
${{\mathbf{A}}}=\left[ {\begin{array}{*{20}{c}}
{{\mathbf{X}}}& \xi({{\mathbf{X}}}{{\mathbf{W}}_{{{h}}}}+{{\bm{\beta }}_{{{h}}}})
\end{array}} \right]$.
Then the expanded input matrix ${\mathbf{A}}$ is applied to compute the output by
${\mathbf{\hat{Y}}}={{\mathbf{A}}}{{\mathbf{W}}}$,
 where
  ${{\mathbf{W}}}$ is the output weight matrix.

  In RVFLNN,   it is only required to train
    the output weights ${{\mathbf{W}}}$  by solving the
 problem
$\mathbf{Y}={{\mathbf{A}}}{{\mathbf{W}}}$,
where ${{\mathbf{Y}}}$ denotes the  labels.
For that problem,
the
least-square solution~\cite{27_ref_BL_trans_paper}
is
 computed
 by
\begin{equation}\label{W2AinvY989565}
\mathbf{W}=\mathbf{A}_{{}}^{+ }\mathbf{Y},
\end{equation}
which is equal to the generalized inverse solution based on
the pseudoinverse (i.e., the generalized inverse)
 \begin{equation}\label{Ainv2AtAinvAt9096}
\mathbf{A}_{{}}^{+}={(\mathbf{A}_{{}}^{T}\mathbf{A})^{-1}}\mathbf{A}_{{}}^{T}.
 \end{equation}

 \subsection{Broad Learning Model}\label{sec2subsec1}

 Unlike the traditional RVFLNN that enhances the input data ${{\mathbf{X}}}$ directly,
 the BLS transfers  the original input data ${{\mathbf{X}}}$   into the mapped features in the feature nodes,
 and then enhances the feature nodes into  the enhancement
nodes. Finally, the connections of all the feature  and enhancement
nodes are fed into the output.


The BLS projects
 the input data $\mathbf{X}$
  by
\begin{equation}\label{Z2PhyXWb985498}
{{\mathbf{Z}}_{i}}=\phi (\mathbf{X}{{\mathbf{W}}_{{{e}_{i}}}}+{{\bm{\beta }}_{{{e}_{i}}}})
\end{equation}
to obtain the $i$-th group of feature nodes ${{\mathbf{Z}}_{i}}$,
while  the  weights ${{\mathbf{W}}_{{{e}_{i}}}}$ and the biases ${{\bm{\beta }}_{{{e}_{i}}}}$ in (\ref{Z2PhyXWb985498}) are
  randomly generated and then fine-tuned
   by applying the linear inverse problem~\cite{BL_trans_paper}.  All the $n$ groups of feature nodes
  \begin{equation}\label{Z2Z1Z2320980esd}
{{\mathbf{Z}}^{n}}\equiv \left[ \begin{matrix}
   {{\mathbf{Z}}_{1}} & \cdots  & {{\mathbf{Z}}_n}  \\
\end{matrix} \right]
 \end{equation}
  are enhanced  to obtain the
$j$-th group of enhancement nodes ${{\mathbf{H}}_{j}}$
  by
\begin{equation}\label{HjipsenZjWbelta09885}
{{\mathbf{H}}_{j}}=\xi ({{\mathbf{Z}}^{n}}{{\mathbf{W}}_{{{h}_{j}}}}+{{\bm{\beta }}_{{{h}_{j}}}}),
\end{equation}
where ${{\mathbf{W}}_{{{h}_{j}}}}$ and
    ${{\bm{\beta }}_{{{h}_{j}}}}$ are random.  Then all the $m$ groups of enhancement nodes are written as
\begin{equation}\label{Hj2H1Hj9859348}
{{\mathbf{H}}^{m}}\equiv \left[ {{\mathbf{H}}_{1}},\cdots, {{\mathbf{H}}_{m}} \right].
\end{equation}

The expanded input matrix, which  consists of all
 the $n$ groups of feature nodes and all the $m$ groups of enhancement nodes, is denoted as
\begin{equation}\label{Anm2ZH3209kds32}
{{\mathbf{A}}_n^{m}}=\left[ {\begin{array}{*{20}{c}}
{{\mathbf{Z}}^{n}}&{{\mathbf{H}}^{m}}
\end{array}} \right]
 \end{equation}
 with
  the subscript $n$  and the superscript $m$ of ${{\mathbf{A}}_n^{m}}$ to be
 the numbers of feature and enhancement node groups, respectively.
The connections of all the feature nodes ${{\mathbf{Z}}^{n}}$
 and all the enhancement nodes ${{\mathbf{H}}^{m}}$
 are fed into the output by
 \begin{equation}\label{}
\mathbf{\hat{Y}}={{\mathbf{A}}_n^{m}}{{\mathbf{W}}_n^{m}},
 \end{equation}
where the
output
 weights 
 \begin{equation}\label{}
{{\mathbf{W}}_n^{m}}=({{\mathbf{A}}_n^{m}})^{+}\mathbf{Y}
 \end{equation}
are computed from the ridge regression approximation of the pseudoinverse, i.e.,
 \begin{equation}\label{AinvLimNumda0AAiA1221}
\mathbf{A}_{{}}^{+ }=\underset{\lambda \to 0}{\mathop{\lim }}\,{(\mathbf{A}_{{}}^{T}\mathbf{A}+\lambda \mathbf{I})^{-1}}\mathbf{A}_{{}}^{T}.
\end{equation}


\subsection{Incremental Learning for Added Inputs}\label{sec2subsec2}

The BLS includes the incremental learning for the additional input training samples.
When  encountering new input samples with the corresponding output labels,
the modeled BLS can be remodeled in an
incremental way without a complete retraining process.
It updates the output weights incrementally, without retraining the whole network from the beginning.

Denote the additional input training samples  as ${{\mathbf{X}}_{a}}$.
 The corresponding additional samples
 in the feature and
enhancement nodes
can be represented as the
matrix  ${\bf{A}}_x$.
Then the expanded input matrix
${\bf{A}}_n^m$
should be updated into
\begin{equation}\label{AxInputIncrease31232USE79547}
{}^x{\bf{A}}_n^m = \left[ \begin{array}{l}
{\bf{A}}_n^m\\
{\bf{A}}_x^{}
\end{array} \right],
\end{equation}
which can be regarded as  a  matrix partitioned along the rows.
Accordingly, the output weights
\begin{equation}\label{Wj2ZiHjY084379}
{\bf{W}}_n^m={{({\bf{A}}_n^m)}^{+}}\mathbf{Y}
\end{equation}
 should be updated into
\begin{equation}\label{xWnm2xAnmYYa3281}
{}^{x}\mathbf{W}_{n}^{m}={{({}^{x}\mathbf{A}_{n}^{m})}^{+}}\left[ \begin{matrix}
   \mathbf{Y}  \\
   {{\mathbf{Y}}_{a}}  \\
\end{matrix} \right],
\end{equation}
where ${{\mathbf{Y}}}$  and ${{\mathbf{Y}}_{a}}$ are the output labels corresponding to
the input ${{\mathbf{X}}}$ and
the additional input  ${{\mathbf{X}}_{a}}$, respectively.

 In \cite{BL_trans_paper}, 
 the pseudoinverse of ${}^x{\bf{A}}_n^m$
 is computed by
  \begin{equation}\label{xAbar2AbarBtDtBt4132OLD984}
{({}^x{\bf{A}}_n^m)^ + }=\left[ \begin{matrix}
   {({\bf{A}}_n^m)^{+ }}-{{\mathbf{B}}}{{\mathbf{D}}^{T}} & {{\mathbf{B}}}  \\
\end{matrix} \right],
\end{equation}

where

\begin{subequations}{\label{B_Matrix_defOriginal3241}}
\begin{numcases}
{  {{\bf{B}}} = }
{ {{\bf{C}}^ + } \quad \quad \quad \quad \quad \quad \quad \quad \quad \quad  \  \  \text{if} \ {\bf{C}} \ne {\bf{0}}}   &  \label{B_Matrix_defOriginal3241a} \\
{{({\bf{A}}_n^m)}^{+}} {{\bf{D}}} {{({\bf{I}} + {{\bf{D}}^T}{\bf{D}})}^{ - 1}} \quad \quad   \ \text{if}\  {\bf{C}} ={\bf{0}},   &   \label{B_Matrix_defOriginal3241b}
\end{numcases}
\end{subequations}
 \begin{equation}\label{CAxDtA413124OLD984}
{\bf{C}} = {\bf{A}}_x^{T} - ({\bf{A}}_n^m)^T {{\bf{D}}},
 \end{equation}
 and
\begin{equation}\label{DtAxAmnInv324141OLD984}
{{\mathbf{D}}^{T}}=\mathbf{A}_{x}^{{}} {({\bf{A}}_n^m)^{+ }}.
\end{equation}
Moreover, the corresponding output weights ${}^{x}\mathbf{W}_{n}^{m}$ are computed by
\begin{equation}\label{xWnm2WnmBAW894353}
{}^x{\bf{W}}_n^m = {\bf{W}}_n^m + {{\bf{B}}}\left( {{{\bf{Y}}_a} - {\bf{A}}_x^{}{\bf{W}}_n^m} \right).
\end{equation}

\section{Proposed Efficient Implementation  for a Step in the Pseudoinverse Computation}\label{sec3}

In this section, we propose an efficient implementation  of  (\ref{B_Matrix_defOriginal3241}), which is a step in the pseudoinverse computation of
 the  partitioned matrix (\ref{AxInputIncrease31232USE79547}).

\subsection{The Proposed Efficient Implementation of a Step in the Pseudoinverse Computation for  BLS}



Let us assume that ${\bf{A}}_n^m$ and ${\bf{A}}_x$  in  (\ref{AxInputIncrease31232USE79547}) are
 $l \times k$ and $q \times k$, respectively, where
  $l$, $q$ and $k$  denote the  number of training samples, the  number of additional training samples and the total number of  nodes, respectively.
Usually it is impossible to observe ${\bf{A}}_n^m$ that is rank deficient~\cite[the last but one paragraph in  pp. 64]{27_ref_BL_trans_paper}.
Then
as in \cite{BL_trans_paper},
 in this paper   we focus on the usual case where the $l \times k$ matrix ${\bf{A}}_n^m$ is of full rank and  $l > k$,
i.e.,
 the   training samples are more
than  the total  nodes~\footnote{Equation (3)  in \cite{BL_trans_paper}
is ${\bf{A}}_{}^ +  = \mathop {\lim }\limits_{\lambda  \to 0} {\mkern 1mu} {(\lambda {\bf{I}} + {\bf{A}}_{}^T{\bf{A}})^{ - 1}}{\bf{A}}_{}^T$, i.e., the left inverse  ${\bf{A}}_{}^ +  = {({\bf{A}}_{}^T{\bf{A}})^{ - 1}}{\bf{A}}_{}^T{\mkern 1mu} $, which assumes the $l \times k$ matrix ${\bf{A}}$ satisfies $l \ge k$ when ${\bf{A}}$ is of  full rank. Moreover, in  Tables
 \Rmnum{5} and \Rmnum{6}
  of \cite{BL_trans_paper} where the simulations include the increment of input pattern, $l \ge 2k$ is satisfied in most cases.}.
Accordingly,
 ${\bf{C}} ={\bf{0}}$  can be concluded, since~\cite[the first paragraph in  pp. 64]{27_ref_BL_trans_paper}
   the $l \times k$ matrix ${\bf{A}}_n^m$ is of full rank and  $l>k$.




In what follows,
we focus on the usual case of ${\bf{C}} ={\bf{0}}$, and then we only need to consider (\ref{B_Matrix_defOriginal3241b}) in (\ref{B_Matrix_defOriginal3241}), into which
we substitute
    (\ref{DtAxAmnInv324141OLD984})
  to obtain
  \begin{equation}\label{B2ADIAADtr65}
{{\bf{B}}}=  {{({\bf{A}}_n^m)}^{+}} {{\bf{D}}} {{({\bf{I}} + {\bf{A}}_x^{}{({\bf{A}}_n^m)^ + } {\bf{D}})}^{ - 1}}.
\end{equation}
Then we write (\ref{B2ADIAADtr65}) as
\begin{equation}\label{DaddDHinvD492573Row}
{{\bf{B}}} = {\bf{\bar D}}{{({\bf{I}} + {\bf{A}}_x{\bf{\bar D}})}^{ - 1}}
\end{equation}
where
\begin{equation}\label{D_bar_def_432ge32}
{\bf{\bar D}} = {{({\bf{A}}_n^m)}^{+}} {{\bf{D}}}.
\end{equation}
  From (\ref{DaddDHinvD492573Row}),
  we deduce
\begin{equation}\label{B2IDAD45982}
{{\bf{B}}} = {{({\bf{I}} + {\bf{\bar D}} {\bf{A}}_x)}^{ - 1}}{\bf{\bar D}}
\end{equation}
  finally, by  utilizing  the inverse of a sum of matrices described by equation (20) in \cite{InverseSumofMatrix8312},
  i.e.,
  \begin{equation}\label{equ20ref8}
{{({\bf{I}} + {{\bf{P}}}{\bf{Q}})}^{ - 1}}{{\bf{P}}} ={{\bf{P}}} {{({\bf{I}} +{\bf{Q}} {{\bf{P}}})}^{ - 1}}.
\end{equation}


${\bf{A}}_x$ is a $q \times k$ matrix, while ${\bf{\bar D}}$ is a
$k \times q$  matrix. Then it can be seen that
  (\ref{B2IDAD45982}) with a $k \times k$ matrix inverse   is usually more
efficient than (\ref{DaddDHinvD492573Row})
and
(\ref{B_Matrix_defOriginal3241b})
 with a $q \times q$
matrix
 inverse when $q > k$, while the latter two equations are
 usually more
efficient than the former
  when $q < k$.  Moreover, it can be seen that
   the computational complexity of ${\bf{A}}_x{\bf{\bar D}}$  in  (\ref{DaddDHinvD492573Row})  is   lower
 than that of ${{\bf{D}}^T}{\bf{D}}$ in
 (\ref{B_Matrix_defOriginal3241b}),
 since we focus on the usual case of $l > k$, as mentioned above.
 Accordingly,  we had better use (\ref{DaddDHinvD492573Row}) instead of
  (\ref{B_Matrix_defOriginal3241b})  when $q < k$.
   Finally, we  
   replace  (\ref{B_Matrix_defOriginal3241b}) with
     (\ref{DaddDHinvD492573Row}) and (\ref{B2IDAD45982}), to improve  
     (\ref{B_Matrix_defOriginal3241})
      into%
\begin{flushleft}
\begin{subequations}
 \begin{numcases}
{  {{\bf{B}}} = }
{ {{\bf{C}}^ + } \quad \quad \quad \quad \quad \quad \quad \quad \  \,  \text{if} \ {\bf{C}} \ne {\bf{0}}}   &  \label{B_Matrix_def2bSimplCne0} \\%
{\bf{\bar D}}{{({\bf{I}} + {\bf{A}}_x{\bf{\bar D}})}^{ - 1}} \quad \quad \quad  \ \text{if}\  {\bf{C}} ={\bf{0}} \And  q \le k   &   \label{B_Matrix_def2bSimplebbb}\\
{{({\bf{I}} + {\bf{\bar D}} {\bf{A}}_x)}^{ - 1}}{\bf{\bar D}} \quad \quad \quad  \ \text{if}\ {\bf{C}} ={\bf{0}} \And  q > k,   &  \label{B_Matrix_def2bSimpleaaa}%
\end{numcases}
{\label{B_Matrix_defSimple3141}}%
\end{subequations}
\end{flushleft}
where ${\bf{\bar D}}$ is computed by (\ref{D_bar_def_432ge32}).

 \subsection{Construction Model and Learning Procedure of the Presented Implementations for the BLS on  Added Inputs}


  \textbf{Algorithm 1} and \textbf{Algorithm 2}
 summarize the construction model and learning procedure of
the original implementation for the  BLS on added inputs~\cite{BL_trans_paper}.
\textbf{Algorithm 1} introduces the initialization phase, which utilizes
(\ref{Z2PhyXWb985498})-(\ref{Anm2ZH3209kds32}).
Then
\textbf{Algorithm 2}  describes the computation of   output weights and the dynamical increment of inputs,
which utilizes (\ref{W2AinvY989565})-(\ref{Anm2ZH3209kds32}) and (\ref{xAbar2AbarBtDtBt4132OLD984})-(\ref{xWnm2WnmBAW894353}).

\begin{algorithm}
\caption{:~\bf The Broad Learning Algorithm:  Computation of the Initial Expanded Input Matrix}
\begin{algorithmic}[1]
\Require  Input data ${{\bf{X}}}$
\Ensure  Expanded input matrix ${\bf{A}}_n^m$
\For{$i=1:n$}
\State Fine-tune random ${{\bf{W}}_{{{e}_{i}}}}$ and  ${{\bm{\beta }}_{{{e}_{i}}}}$;
\State Compute ${{\bf{ Z}}_{i}}=\phi ({{\bf{X}}} {{\bf{W}}_{{{e}_{i}}}}+{{\bm{\beta }}_{{{e}_{i}}}})$;
\EndFor
\State Concatenate the feature nodes into ${{\bf{ Z}}^{n}}\equiv \left[ \begin{matrix}
   {{\bf{ Z}}_{1}} & \cdots  & {{\bf{ Z}}_n}  \\
\end{matrix} \right]$;
\For{$j=1:m$}
\State Random   ${{\bf{W}}_{{{h}_{j}}}}$ and
    ${{\bm{\beta }}_{{{h}_{j}}}}$;
\State Compute ${{\bf{ H}}_{j}}=\xi ({{\bf{ Z}}^{n}}{{\bf{W}}_{{{h}_{j}}}}+{{\bm{\beta }}_{{{h}_{j}}}})$;
\EndFor
\State   Set the enhancement nodes group 
${{\bf{ H}}^{m}}\equiv \left[ {{\bf{ H}}_{1}},\cdots,{{\bf{ H}}_{m}} \right]$;
\State Set ${{\bf{ A}}_n^{m}}= \left[ {\begin{array}{*{20}{c}}
{{\bf{ Z}}^{n}}&{{\bf{ H}}^{m}}
\end{array}} \right]$;
\end{algorithmic}
\end{algorithm}

\begin{algorithm}
\caption{:~\bf The Original Implementation to Compute Output Weights and Add Input Samples for BLS}
\begin{algorithmic}[1]
\Require Expanded input matrix ${{\mathbf{ A}}_n^{m}}$, labels $\mathbf{Y}$, additional input training samples  ${{\mathbf{X}}_{a}}$
and labels ${{\mathbf{Y}}_{a}}$
\Ensure Output weights ${\bf{W}}_n^m$
\State Compute $({{\mathbf{ A}}_n^{m}})^{+ }=\underset{\lambda \to 0}{\mathop{\lim }}\,{\left(({{\mathbf{ A}}_n^{m}})^{T}{{\mathbf{ A}}_n^{m}}+\lambda \mathbf{I}\right)^{-1}}({{\mathbf{ A}}_n^{m}})^{T}$; \State Compute ${\bf{W}}_n^m=({{\mathbf{ A}}_n^{m}})^{+ }\mathbf{Y}$;
\While{\emph{The target training error  is not reached}}
\State Use   ${{\mathbf{X}}_{a}}$ to get
 ${\bf{A}}_x \in {\Re ^{q \times k}}$
by  (\ref{Z2PhyXWb985498})-(\ref{Anm2ZH3209kds32});
\State  
Compute ${{\mathbf{D}}^{T}}=\mathbf{A}_{x}^{{}} {({\bf{A}}_n^m)^{+ }}$;
\State Compute  ${\bf{C}} = {\bf{A}}_x^{T} - ({\bf{A}}_n^m)^T {{\bf{D}}}$ or assume  ${\bf{C}} ={\bf{0}}$;
\State Compute \begin{numcases}
{  {{\bf{B}}} = }
{ {{\bf{C}}^ + } \quad \quad \quad \quad \quad \quad \quad \quad \quad \quad  \  \  \text{if} \ {\bf{C}} \ne {\bf{0}}}   &  \notag \\
{{({\bf{A}}_n^m)}^{+}} {{\bf{D}}} {{({\bf{I}} + {{\bf{D}}^T}{\bf{D}})}^{ - 1}} \quad \quad   \ \text{if}\  {\bf{C}} ={\bf{0}};   &   \notag
\end{numcases}
\State  Compute ${({}^x{\bf{A}}_n^m)^ + }=\left[ \begin{matrix}
   {({\bf{A}}_n^m)^{+ }}-{{\mathbf{B}}}{{\mathbf{D}}^{T}} & {{\mathbf{B}}}  \\
\end{matrix} \right]$;
\State  
Compute ${}^x{\bf{W}}_n^m = {\bf{W}}_n^m + {{\bf{B}}}\left( {{{\bf{Y}}_a} - {\bf{A}}_x^{}{\bf{W}}_n^m} \right)$;
\State   $ l \Leftarrow l+q $;
\State   $ {({}^x{\bf{A}}_n^m)^ + } \Leftarrow {({\bf{A}}_n^m)^{+ }} $;
\State   $ {\bf{W}}_n^m \Leftarrow {}^x{\bf{W}}_n^m $;
\EndWhile
\end{algorithmic}
\end{algorithm}

The original implementation for the  BLS on added inputs
computes  ${{\bf{B}}}$ by
 (\ref{B_Matrix_defOriginal3241}), while the proposed efficient
 implementation for the  BLS on added inputs
  computes  ${{\bf{B}}}$ by
 (\ref{D_bar_def_432ge32}) and (\ref{B_Matrix_defSimple3141}).
 For the proposed efficient
 implementation, \textbf{Algorithm 3}  describes
 the computation of   output weights and the dynamical increment of inputs,
which utilizes (\ref{W2AinvY989565})-(\ref{Anm2ZH3209kds32}),  (\ref{xAbar2AbarBtDtBt4132OLD984}), (\ref{CAxDtA413124OLD984})-(\ref{xWnm2WnmBAW894353}), (\ref{D_bar_def_432ge32}) and (\ref{B_Matrix_defSimple3141}).

\begin{algorithm}
\caption{:~\bf The Proposed Efficient Implementation to Compute Output Weights and Add Input Samples for BLS}
\begin{algorithmic}[1]
\Require Expanded input matrix ${{\mathbf{ A}}_n^{m}}$, labels $\mathbf{Y}$, additional input training samples  ${{\mathbf{X}}_{a}}$
and labels ${{\mathbf{Y}}_{a}}$
\Ensure Output weights ${\bf{W}}_n^m$ 
\State Compute $({{\mathbf{ A}}_n^{m}})^{+ }=\underset{\lambda \to 0}{\mathop{\lim }}\,{\left(({{\mathbf{ A}}_n^{m}})^{T}{{\mathbf{ A}}_n^{m}}+\lambda \mathbf{I}\right)^{-1}}({{\mathbf{ A}}_n^{m}})^{T}$; \State Compute ${\bf{W}}_n^m=({{\mathbf{ A}}_n^{m}})^{+ }\mathbf{Y}$;
\While{\emph{The target training error  is not reached}}
\State Use  ${{\mathbf{X}}_{a}}$ to get
${\bf{A}}_x \in {\Re ^{q \times k}}$
by  (\ref{Z2PhyXWb985498})-(\ref{Anm2ZH3209kds32});
\State Compute ${{\mathbf{D}}^{T}}=\mathbf{A}_{x}^{{}} {({\bf{A}}_n^m)^{+ }}$;
\State Compute ${\bf{\bar D}} = {{({\bf{A}}_n^m)}^{+}} {{\bf{D}}}$;
\State Compute ${\bf{C}} = {\bf{A}}_x^{T} - ({\bf{A}}_n^m)^T {{\bf{D}}}$  or assume  ${\bf{C}} ={\bf{0}}$;
\State Compute
 \begin{numcases}
{  {{\bf{B}}} = }
{ {{\bf{C}}^ + } \quad \quad \quad \quad \quad \quad \quad \quad \  \,  \text{if} \ {\bf{C}} \ne {\bf{0}}}   &  \notag \\
{\bf{\bar D}}{{({\bf{I}} + {\bf{A}}_x{\bf{\bar D}})}^{ - 1}} \quad \quad \quad  \ \text{if}\  {\bf{C}} ={\bf{0}} \And  q \le k   &   \notag \\
{{({\bf{I}} + {\bf{\bar D}} {\bf{A}}_x)}^{ - 1}}{\bf{\bar D}} \quad \quad \quad  \ \text{if}\ {\bf{C}} ={\bf{0}} \And  q > k;   &  \notag
\end{numcases}
\State Compute
${({}^x{\bf{A}}_n^m)^ + }=\left[ \begin{matrix}
   {({\bf{A}}_n^m)^{+ }}-{{\mathbf{B}}}{{\mathbf{D}}^{T}} & {{\mathbf{B}}}  \\
\end{matrix} \right]$;
\State Compute ${}^x{\bf{W}}_n^m = {\bf{W}}_n^m + {{\bf{B}}}\left( {{{\bf{Y}}_a} - {\bf{A}}_x^{}{\bf{W}}_n^m} \right)$;
\State   $ l \Leftarrow l+q $;
\State   $ {({}^x{\bf{A}}_n^m)^ + } \Leftarrow {({\bf{A}}_n^m)^{+ }} $;
\State   $ {\bf{W}}_n^m \Leftarrow {}^x{\bf{W}}_n^m $;
\EndWhile
\end{algorithmic}
\end{algorithm}

\subsection{Complexity  Comparison between  the Original  and Proposed  BLS Implementations}

As mentioned above,
 in the usual case where  ${\bf{A}}_n^m \in {\Re ^{l \times k}}$ is of full rank and  $l > k$,
  ${\bf{C}} ={\bf{0}}$  is satisfied in
  (\ref{B_Matrix_defOriginal3241}) and
 (\ref{B_Matrix_defSimple3141}).
Thus, in this subsection we assume that ${\bf{C}} ={\bf{0}}$  is satisfied, and
  compute the expected flops
(floating-point operations)
 of
 the BLS using the proposed implementation (\ref{B_Matrix_def2bSimpleaaa}) or
 (\ref{B_Matrix_def2bSimplebbb})
 and
the BLS using the original implementation (\ref{B_Matrix_defOriginal3241b}).

It can easily be seen that
 $l q (2 k - 1) \approx 2 l  k q$ flops
 are required
 to multiply
 a $l \times k$ matrix by a $k \times q$ matrix, and
  $l  k=0(l  k q)$  flops
 are required
 to sum two matrices in size $l \times k$.
The  inv function in Matlab~\cite{Matlab_inv_function_introduce}
includes the computation of the ${\mathbf{LDL}}^T$ factors of
the  Hermitian matrix $\mathbf{X}$, and the subsequent invert-and-multiply step to
invert the factors and multiply the inverses.
For the $k \times k$ Hermitian matrix $\mathbf{X}$,
the computation of the ${\mathbf{LDL}}^T$ factors and the subsequent invert-and-multiply step
require $\frac{1}{3} k^3$ flops~\cite{Matrix_Computations_book}
 and $\frac{2}{3} k^3$ flops~\cite{BackSubsitutionCholeskyflops},
 respectively.
 Thus it totally requires $k^3$ flops  to compute the inverse of the Hermitian matrix $\mathbf{X}$.
 Moreover, the  inv function in Matlab  computes the inverse of the $k \times k$ non-Hermitian matrix $\mathbf{X}$ by
 the $\mathbf{LU}$ factorization and the subsequent invert-and-multiply step, which
 totally require $2 k^3$ flops.

Table~\ref{tab1}
gives a step-by-step comparison
between
the flops of
the proposed
implementation (\ref{B_Matrix_def2bSimpleaaa}) with (\ref{D_bar_def_432ge32}) for the case of $q > k$
and those of the original implementation  (\ref{B_Matrix_defOriginal3241b}).
 From Table~\ref{tab1}, it can be seen that
the proposed
 (\ref{B_Matrix_def2bSimpleaaa})
 speeds up the matrix inversion operation and two relevant matrix multiplication operations in
 the original   (\ref{B_Matrix_defOriginal3241b}), by the speed-up  factors
 of $\frac{1}{2}(q/k)^3$, $ql/(2 k^2)$ and $q/k$, respectively, where $q >  k$ and $l > k$, as mentioned previously.
Moreover,
  it can be seen from Table~\ref{tab1} that the total flops
  of  the original implementation  (\ref{B_Matrix_defOriginal3241b}) and the proposed
implementation (\ref{B_Matrix_def2bSimpleaaa}) with (\ref{D_bar_def_432ge32}) are
    \begin{equation}\label{flopsExisting}
2 qkl +q^2 l +  q^3 + 2 q^2 k
 \end{equation}
and
    \begin{equation}\label{flopsProposedBest}
2 qkl + 4k^2 q + 2k^3,
 \end{equation}
 respectively.

 \begin{table}[!t]
\begin{center}
\begin{minipage}{\textwidth}
\scriptsize
\renewcommand{\arraystretch}{1.8} 
\newcommand{\tabincell}[2]{\begin{tabular}{@{}#1@{}}#2\end{tabular}}
\caption{Comparison of Flops between  the Proposed
Implementation (\ref{B_Matrix_def2bSimpleaaa}) with (\ref{D_bar_def_432ge32}) for the Case of $q > k$ and the Original Implementation  (\ref{B_Matrix_defOriginal3241b})} \label{tab1} \centering
\begin{tabular}{|c|c|c|c|}
\multicolumn{2}{|c|}{{\bfseries   \tabincell{c}{Proposed (\ref{B_Matrix_def2bSimpleaaa}) with  (\ref{D_bar_def_432ge32}) for $q > k$}}}  &  \multicolumn{2}{c|}{{\bfseries   \tabincell{c}{Original (\ref{B_Matrix_defOriginal3241b})}}}    \\
{\bfseries   \tabincell{c}{Operation}}     & {\bfseries   \tabincell{c}{Flops}}  & {\bfseries   \tabincell{c}{Operation}}     & {\bfseries   \tabincell{c}{Flops}}  \\
\hline
\hline
 ${{({\bf{A}}_n^m)}^{+}} \times {{\bf{D}}}$ (i.e., (\ref{D_bar_def_432ge32})) & $2 qkl$   &  ${{({\bf{A}}_n^m)}^{+}} \times {{\bf{D}}}$  & $2 qkl$   \\
 \hline
     ${{({\bf{I}} + {\bf{\bar D}} {\bf{A}}_x)}^{ - 1}}$ (in (\ref{B_Matrix_def2bSimpleaaa})) &   $2 k^3$
       &  ${{({\bf{I}} + {{\bf{D}}^T}{\bf{D}})}^{ - 1}}$  & $q^3$ \\
\hline
${\bf{\bar D}}  \times  {\bf{A}}_x$ (in (\ref{B_Matrix_def2bSimpleaaa})) & $2 k^2 q$  & ${{\bf{D}}^T} \times {\bf{D}}$  & $q^2 l$\\
\hline
 \tabincell{c}{${{({\bf{I}} + {\bf{\bar D}} {\bf{A}}_x)}^{ - 1}} \times {\bf{\bar D}}$ (in (\ref{B_Matrix_def2bSimpleaaa}))}   & $2 k^2 q$
  &  \tabincell{c}{$\left( {{({\bf{A}}_n^m)}^{+}} {{\bf{D}}} \right) \times {{({\bf{I}} + {{\bf{D}}^T}{\bf{D}})}^{ - 1}}$}  & $2 q^2 k$
  \\
\hline
\end{tabular}
\end{minipage}
\end{center}
\end{table}

  The proposed
implementation (\ref{B_Matrix_def2bSimplebbb}) with (\ref{D_bar_def_432ge32}) for the case of $q \le  k$
only replaces ${{\bf{D}}^T}  {\bf{D}}$ in (\ref{B_Matrix_defOriginal3241b})
with ${\bf{A}}_x   {\bf{\bar D}}$. The flops to compute the Hermitian
${\bf{A}}_x {\bf{\bar D}}$ are $q^2 k$,
and as shown in Table~\ref{tab1}, the flops to compute the Hermitian ${{\bf{D}}^T}{\bf{D}}$
are $q^2 l$. Thus  the proposed
 (\ref{B_Matrix_def2bSimplebbb}) speeds up the matrix multiplication operation ${{\bf{D}}^T} \times {\bf{D}}$  in
 the original   (\ref{B_Matrix_defOriginal3241b}) by the speed-up  factor of
\begin{equation}\label{flopsProposedSaveSmall}
(q^2 l)/ (q^2 k)= l/k,
\end{equation}
 where
  $l > k$, as mentioned previously.

The dominant flops
 of
(\ref{DtAxAmnInv324141OLD984}),
(\ref{CAxDtA413124OLD984})
and
(\ref{xAbar2AbarBtDtBt4132OLD984})
 are $2qkl$, $2qkl$ and $2qkl$, respectively, and the dominant flops to update ${{}^{x}\mathbf{W}_{n}^{m}}$ by
 (\ref{xWnm2WnmBAW894353})
  are $2cqk+2cqk=4cqk$, where $c$ denotes the number of output nodes. Then the total flops of
  (\ref{DtAxAmnInv324141OLD984}),
(\ref{CAxDtA413124OLD984}),
(\ref{xAbar2AbarBtDtBt4132OLD984})
and
(\ref{xWnm2WnmBAW894353}) are
    \begin{equation}\label{totalflops4equ}
2qkl+2qkl+2qkl+4cqk=6qkl+4cqk.
 \end{equation}

 From
 (\ref{flopsExisting})-(\ref{totalflops4equ}),
we can deduce that in each update for the incremental learning on added inputs,
  the total flops
  of  the  BLS using the original implementation (\ref{B_Matrix_defOriginal3241b}),  the BLS using the proposed
implementation (\ref{B_Matrix_def2bSimplebbb}) for $q \le  k$,   and the BLS using the proposed
implementation (\ref{B_Matrix_def2bSimpleaaa}) for $q >  k$
are
 \begin{equation}\label{flopsAllexisting}
8 qkl + 4cqk + q^2 l +  q^3 + 2 q^2 k,
 \end{equation}
    \begin{equation}\label{flopsAllmine111}
8 qkl + 4cqk + q^2 k +  q^3 + 2 q^2 k,
 \end{equation}
and
    \begin{equation}\label{flopsAllmine222}
8 qkl + 4cqk + 4k^2 q + 2k^3,
 \end{equation}
 respectively.

\section{Numerical Experiments}\label{sec4}

In this section, we carry out numerical experiments
on MATLAB software platform under a Microsoft-Windows Server with  $128$ GB of RAM,
to evaluate the testing accuracies and training time
of
 the BLS using the original implementation (i.e., (\ref{B_Matrix_defOriginal3241b}))
and
the BLS using the proposed efficient implementation (i.e.,  (\ref{B_Matrix_def2bSimpleaaa}) or
 (\ref{B_Matrix_def2bSimplebbb})), which have been described in \textbf{Algorithm 2} and \textbf{Algorithm 3},
 respectively.
  In the simulations,
   the tansig function is  adopted for the
enhancement nodes, where the weights
    ${{\mathbf{W}}_{{{h}_{j}}}}$ and
   the biases
    ${{\mathbf{\beta }}_{{{h}_{j}}}}$ ($j=1,2,\cdots, m$)
    are drawn from the
standard uniform distributions on the interval $\left[ {\begin{array}{*{20}{c}}
{{\rm{ - }}1}&1
\end{array}} \right]$.

\begin{table*}[!t]
 \begin{center}
\begin{minipage}{\textwidth}
 \scriptsize
\renewcommand{\arraystretch}{1.3}
\setlength\tabcolsep{2pt}%
\newcommand{\tabincell}[2]{\begin{tabular}{@{}#1@{}}#2\end{tabular}}
\caption{Testing Accuracies and Training Time of the  Implementations for
BLS on MNIST Dataset with $k=5100$ nodes (i.e., $5000$ enhancement and $100$ feature nodes)
and $q=10000(>k)$ added training samples in each update} \label{tab2}
 \centering
\begin{tabular}{|c||c|c  c| c |c c|c|}
\hline
\multirow{4}*{ {\bfseries   \tabincell{c}{Number \\ of \\  Input \\ Patterns}}}     &  \multirow{4}*{ {\bfseries   \tabincell{c}{Testing \\ Accuracy \\ ($\% $)}}}      &\multicolumn{3}{c|}{{\bfseries  Each Additional }}       &\multicolumn{3}{c|}{{\bfseries  Accumulative   }}  \\
   &       &\multicolumn{3}{c|}{{\bfseries  Training    }}
       &\multicolumn{3}{c|}{{\bfseries  Training  }}  \\
   &       &\multicolumn{3}{c|}{{\bfseries   Time   }}
       &\multicolumn{3}{c|}{{\bfseries  Time  }}  \\
   \cdashline{3-8}
      &  &Orig.        &Impr.       & {{\bfseries   Speedups}}   &Orig.        &Impr.      &{{\bfseries Speedups}}      \\
\hline
  \bfseries 10000   & 97.57     ($\pm$     0.11)  &    4.57     &    4.55     &       &    4.57     &    4.55     &           \\
\hdashline
  \bfseries  $\rightarrow$ 20000   &98.28     ($\pm$     0.07)  &   26.07     &   19.42     &    {\bfseries 1.34}     &   30.64     &   23.97     &    {\bfseries 1.28}  \\
\hdashline
  \bfseries  $\rightarrow$ 30000  &98.45     ($\pm$     0.06)  &   45.31     &   33.39     &   {\bfseries 1.36}     &   75.96     &   57.36     &   {\bfseries 1.32}  \\
\hdashline
 \bfseries  $\rightarrow$ 40000     &98.57     ($\pm$     0.07)  &   63.75     &   46.99     &   {\bfseries 1.36}     &  139.71     &  104.35     &  {\bfseries  1.34} \\
\hdashline
 \bfseries  $\rightarrow$ 50000  &98.59     ($\pm$     0.06)  &   82.67     &   61.04     &   {\bfseries 1.35}     &  222.38     &  165.39     &   {\bfseries 1.34}\\
  \hdashline
\bfseries  $\rightarrow$ 60000    &98.64     ($\pm$     0.06)  &  101.71     &   74.79     &  {\bfseries  1.36}     &  324.09     &  240.18     &  {\bfseries  1.35}\\
\hline
\end{tabular}
\end{minipage}
\end{center}
\end{table*}

Tables~\ref{tab2}-\ref{tab5} give the experimental results
on the Modified National Institute of Standards and Technology (MNIST)
dataset~\cite{61_dataSet}, which includes $70000$ handwritten digital images  partitioned into a training set with $60000$  images
  and a test set with
   $10000$  images.  As
 the  simulations
for Table  \Rmnum{5} in \cite{BL_trans_paper},
the simulations for Table~\ref{tab2}
utilize
the initial
 $l=10000$ training samples to
train the  network
with $k=5100$ nodes that  include  $10 \times 10$ feature nodes and $5000$ enhancement nodes,
and add $q=10000>k$ training samples in each update, till all the
$60000$ training samples are fed. Then the simulations for Table~\ref{tab3}
utilize
the initial
 $l=15000$ training samples
 to train the network
 with $k=5100$ nodes  utilized in
  Table~\ref{tab2},
and  add $q=15000>k$ training samples in each update.
On the other hand, the simulations for Tables~\ref{tab4}  and \ref{tab5},
which also
follow  the network with $k=5100$ nodes  utilized in Table~\ref{tab2},
add
 $q=5000<k$ and $q=3500<k$ training samples in each update, respectively,
  to increase the inputs
   to the final $60000$ after $5$ updates.


\begin{table*}[!t]
 \begin{center}
\begin{minipage}{\textwidth}
 \scriptsize
\renewcommand{\arraystretch}{1.3}
\setlength\tabcolsep{2pt}%
\newcommand{\tabincell}[2]{\begin{tabular}{@{}#1@{}}#2\end{tabular}}
\caption{Testing Accuracies and Training Time of the  Implementations for
BLS on MNIST Dataset with $k=5100$ nodes (i.e., $5000$ enhancement and $100$ feature nodes)
and $q=15000(>k)$ added training samples in each update} \label{tab3} \centering
\begin{tabular}{|c||c|c  c| c |c c|c|}
\hline
\multirow{4}*{ {\bfseries   \tabincell{c}{Number \\ of \\  Input \\ Patterns}}}     &  \multirow{4}*{ {\bfseries   \tabincell{c}{Testing \\ Accuracy \\ ($\% $)}}}      &\multicolumn{3}{c|}{{\bfseries  Each Additional }}       &\multicolumn{3}{c|}{{\bfseries  Accumulative   }}  \\
   &       &\multicolumn{3}{c|}{{\bfseries  Training    }}
       &\multicolumn{3}{c|}{{\bfseries  Training  }}  \\
   &       &\multicolumn{3}{c|}{{\bfseries   Time   }}
       &\multicolumn{3}{c|}{{\bfseries  Time  }}  \\
   \cdashline{3-8}
      &  &Orig.        &Impr.       & {{\bfseries   Speedups}}   &Orig.        &Impr.      &{{\bfseries Speedups}}      \\
\hline
  \bfseries 15000    &   98.14   ($\pm$     0.08  )  &     6.33  &    6.32  &     &    6.33  &    6.32  &        \\
\hdashline
   \bfseries  $\rightarrow$ 30000  &98.49   ($\pm$     0.06  )  &    62.64  &   37.70  &  {\bfseries  1.66}  &   68.97  &   44.02  &   {\bfseries 1.57}    \\
\hdashline
  \bfseries  $\rightarrow$ 45000    & 98.57   ($\pm$     0.06  )  &   108.24  &   68.44  &  {\bfseries  1.58}  &  177.20  &  112.45  &  {\bfseries  1.58}    \\
\hdashline
 \bfseries  $\rightarrow$ 60000   &98.64   ($\pm$     0.05  )  &   153.61  &   98.33  &  {\bfseries  1.56}  &  330.81  &  210.78  &  {\bfseries  1.57}    \\
\hline
\end{tabular}
\end{minipage}
\end{center}
\end{table*}

\begin{table*}[!t]
 \begin{center}
\begin{minipage}{\textwidth}
 \scriptsize
\renewcommand{\arraystretch}{1.3}
\setlength\tabcolsep{2pt}%
\newcommand{\tabincell}[2]{\begin{tabular}{@{}#1@{}}#2\end{tabular}}
\caption{Testing Accuracies and Training Time of the  Implementations for
BLS on MNIST Dataset with $k=5100$ nodes (i.e., $5000$ enhancement and $100$ feature nodes)
and $q=5000(<k)$ added training samples in each update} \label{tab4}
 \centering
\begin{tabular}{|c||c|c  c| c |c c|c|}
\hline
\multirow{4}*{ {\bfseries   \tabincell{c}{Number \\ of \\  Input \\ Patterns}}}     &  \multirow{4}*{ {\bfseries   \tabincell{c}{Testing \\ Accuracy \\ ($\% $)}}}      &\multicolumn{3}{c|}{{\bfseries  Each Additional }}       &\multicolumn{3}{c|}{{\bfseries  Accumulative   }}  \\
   &       &\multicolumn{3}{c|}{{\bfseries  Training    }}
       &\multicolumn{3}{c|}{{\bfseries  Training  }}  \\
   &       &\multicolumn{3}{c|}{{\bfseries   Time   }}
       &\multicolumn{3}{c|}{{\bfseries  Time  }}  \\
   \cdashline{3-8}
      &  &Orig.        &Impr.       & {{\bfseries   Speedups}}   &Orig.        &Impr.      &{{\bfseries Speedups}}      \\
\hline
  \bfseries 35000    & 98.53     ($\pm$     0.06)  &  14.43    &   14.61    &       &   14.43    &   14.61    &    \\
\hdashline
 \bfseries  $\rightarrow$ 40000   &  98.56     ($\pm$     0.05)  & 31.34    &   27.69    &   {\bfseries 1.13}    &   45.77    &   42.30    &  {\bfseries  1.08}  \\
\hdashline
  \bfseries  $\rightarrow$ 45000  & 98.57     ($\pm$     0.05)  & 35.33    &   31.11    & {\bfseries   1.14}    &   81.10    &   73.41    &  {\bfseries  1.10} \\
\hdashline
\bfseries  $\rightarrow$ 50000     & 98.60     ($\pm$     0.05)  & 39.58    &   34.64    &  {\bfseries  1.14}    &  120.68    &  108.04    &   {\bfseries 1.12} \\
\hdashline
 \bfseries  $\rightarrow$ 55000   &98.63     ($\pm$     0.05)  & 44.13    &   38.40    &  {\bfseries  1.15}    &  164.81    &  146.44    &  {\bfseries  1.13} \\
  \hdashline
\bfseries  $\rightarrow$ 60000    &98.64     ($\pm$     0.05) &  47.81    &   41.48    &  {\bfseries  1.15}    &  212.62    &  187.93    &  {\bfseries  1.13} \\
\hline
\end{tabular}
\end{minipage}
\end{center}
\end{table*}

\begin{table*}[!t]
 \begin{center}
\begin{minipage}{\textwidth}
 \scriptsize
\renewcommand{\arraystretch}{1.3}
\setlength\tabcolsep{2pt}%
\newcommand{\tabincell}[2]{\begin{tabular}{@{}#1@{}}#2\end{tabular}}
\caption{Testing Accuracies and Training Time of the  Implementations for
BLS on MNIST Dataset with $k=5100$ nodes (i.e., $5000$ enhancement and $100$ feature nodes)
and $q=3500(<k)$ added training samples in each update} \label{tab5}
 \centering
\begin{tabular}{|c||c|c  c| c |c c|c|}
\hline
\multirow{4}*{ {\bfseries   \tabincell{c}{Number \\ of \\  Input \\ Patterns}}}     &  \multirow{4}*{ {\bfseries   \tabincell{c}{Testing \\ Accuracy \\ ($\% $)}}}      &\multicolumn{3}{c|}{{\bfseries  Each Additional }}       &\multicolumn{3}{c|}{{\bfseries  Accumulative   }}  \\
   &       &\multicolumn{3}{c|}{{\bfseries  Training    }}
       &\multicolumn{3}{c|}{{\bfseries  Training  }}  \\
   &       &\multicolumn{3}{c|}{{\bfseries   Time   }}
       &\multicolumn{3}{c|}{{\bfseries  Time  }}  \\
   \cdashline{3-8}
      &  &Orig.        &Impr.       & {{\bfseries   Speedups}}   &Orig.        &Impr.      &{{\bfseries Speedups}}      \\
\hline
  \bfseries 42500    &  98.57    ($\pm$     0.07  )  &    17.44   &   17.69   &       &   17.44   &   17.69   &         \\
\hdashline
  \bfseries  $\rightarrow$ 46000    &  98.59    ($\pm$     0.06  )  &    25.16   &   22.86   &  {\bfseries   1.10}  &   42.59   &   40.56   &  {\bfseries   1.05}     \\
\hdashline
  \bfseries   $\rightarrow$ 49500  &  98.60   ($\pm$     0.06  )  &    26.75   &   24.41   &  {\bfseries   1.10}  &   69.35   &   64.96   &  {\bfseries   1.07}     \\
\hdashline
 \bfseries  $\rightarrow$ 53000     & 98.63    ($\pm$     0.06  )  &    29.35   &   26.53   &   {\bfseries  1.11}   &   98.7 0  &   91.50  &   {\bfseries  1.08}     \\
\hdashline
 \bfseries  $\rightarrow$ 56500   &98.63    ($\pm$     0.06  )  &    30.78   &   27.86   &  {\bfseries   1.10}  &  129.48   &  119.35   &  {\bfseries   1.08}     \\
  \hdashline
\bfseries  $\rightarrow$ 60000    & 98.65    ($\pm$     0.06  )  &    32.82   &   29.61   &  {\bfseries   1.11}   &  162.3 0  &  148.97   &  {\bfseries   1.09}     \\
\hline
\end{tabular}
\end{minipage}
\end{center}
\end{table*}

 Tables~\ref{tab6}-\ref{tab9} give the experimental results
on the New York University object recognition
benchmark (NORB) dataset~\cite{Norb_dataSet}, which
 consists of $48600$ images
  belonging to five distinct categories: (1) animals; (2) humans;
  (3) airplanes; (4) trucks; and (5) cars.
 Half of these images
  are selected into the training set, and the other half form
 the test set.
The simulations for Tables~\ref{tab6} and \ref{tab7}, which utilize the initial
 $l=9900$ training samples   to
train the  network
with  $k=3500$  nodes that  include  $100 \times 10$ feature nodes
and $2500$ enhancement nodes,
 add $q=7200>k$ and $q=3600>k$ training samples  in each update,
 respectively, till all the
$24300$ training samples are fed.
On the other hand,
the simulations for Tables~\ref{tab8}  and \ref{tab9}, which
train the  network
with  $k=3100$  nodes that  include  $100 \times 10$ feature nodes and $2100$ enhancement nodes,
 add $q=3000<k$ and  $q=2000<k$ training samples
   in each update, respectively,  till all the
$24300$ training samples are fed after 4 updates.

\begin{table*}[!t]
 \begin{center}
\begin{minipage}{\textwidth}
 \scriptsize
\renewcommand{\arraystretch}{1.3}
\setlength\tabcolsep{2pt}%
\newcommand{\tabincell}[2]{\begin{tabular}{@{}#1@{}}#2\end{tabular}}
\caption{Testing Accuracies and Training Time of the  Implementations for
BLS on NORB Dataset with $k=3500$ nodes (i.e., $2500$ enhancement and $1000$ feature nodes)
and $q=7200(>k)$ added training samples in each update} \label{tab6} \centering
\begin{tabular}{|c||c|c  c| c |c c|c|}
\hline
\multirow{4}*{ {\bfseries   \tabincell{c}{Number \\ of \\  Input \\ Patterns}}}     &  \multirow{4}*{ {\bfseries   \tabincell{c}{Testing \\ Accuracy \\ ($\% $)}}}      &\multicolumn{3}{c|}{{\bfseries  Each Additional }}       &\multicolumn{3}{c|}{{\bfseries  Accumulative   }}  \\
   &       &\multicolumn{3}{c|}{{\bfseries  Training    }}
       &\multicolumn{3}{c|}{{\bfseries  Training  }}  \\
   &       &\multicolumn{3}{c|}{{\bfseries   Time   }}
       &\multicolumn{3}{c|}{{\bfseries  Time  }}  \\
   \cdashline{3-8}
      &  &Orig.        &Impr.       & {{\bfseries   Speedups}}   &Orig.        &Impr.      &{{\bfseries Speedups}}      \\
\hline
  \bfseries 9900   &88.78    ($\pm$     0.33  )  &     2.15   &    2.16   &        &    2.15   &    2.16   &          \\
\hdashline
  \bfseries  $\rightarrow$ 17100   &89.27    ($\pm$     0.33  )  &    12.00    &    8.69   &  {\bfseries   1.38}   &   14.16   &   10.86   & {\bfseries 1.30}    \\
\hdashline
  \bfseries  $\rightarrow$ 24300   &89.38    ($\pm$     0.33  )  &    19.11   &   13.72   &  {\bfseries   1.39}   &   33.26   &   24.58   & {\bfseries 1.35 }    \\
\hline
\end{tabular}
\end{minipage}
\end{center}
\end{table*}

\begin{table*}[!t]
 \begin{center}
\begin{minipage}{\textwidth}
 \scriptsize
\renewcommand{\arraystretch}{1.3}
\setlength\tabcolsep{2pt}%
\newcommand{\tabincell}[2]{\begin{tabular}{@{}#1@{}}#2\end{tabular}}
\caption{Testing Accuracies and Training Time of the  Implementations for
BLS on NORB Dataset with $k=3500$ nodes (i.e., $2500$ enhancement and $1000$ feature nodes)
and $q=3600(>k)$ added training samples in each update} \label{tab7} \centering
\begin{tabular}{|c||c|c  c| c |c c|c|}
\hline
\multirow{4}*{ {\bfseries   \tabincell{c}{Number \\ of \\  Input \\ Patterns}}}     &  \multirow{4}*{ {\bfseries   \tabincell{c}{Testing \\ Accuracy \\ ($\% $)}}}      &\multicolumn{3}{c|}{{\bfseries  Each Additional }}       &\multicolumn{3}{c|}{{\bfseries  Accumulative   }}  \\
   &       &\multicolumn{3}{c|}{{\bfseries  Training    }}
       &\multicolumn{3}{c|}{{\bfseries  Training  }}  \\
   &       &\multicolumn{3}{c|}{{\bfseries   Time   }}
       &\multicolumn{3}{c|}{{\bfseries  Time  }}  \\
   \cdashline{3-8}
      &  &Orig.        &Impr.       & {{\bfseries   Speedups}}   &Orig.        &Impr.      &{{\bfseries Speedups}}      \\
\hline
  \bfseries 9900    & 88.79     ($\pm$    0.35)   &    2.23      &    2.22      &          &    2.23      &    2.22      &          \\
\hdashline
\bfseries  $\rightarrow$ 13500   &88.99     ($\pm$    0.35)   &    4.93      &    4.52      &  {\bfseries   1.09}      &    7.15      &    6.75      &  {\bfseries   1.06} \\
\hdashline
  \bfseries  $\rightarrow$ 17100  & 89.28     ($\pm$    0.35)   &    6.37      &    5.78      &   {\bfseries  1.10}      &   13.52      &   12.53      &   {\bfseries  1.08} \\
  \hdashline
\bfseries  $\rightarrow$ 20700     &89.33     ($\pm$    0.36)   &    8.07      &    7.23      &  {\bfseries   1.12}      &   21.59      &   19.76      &   {\bfseries  1.09} \\
  \hdashline
 \bfseries  $\rightarrow$ 24300   &89.38     ($\pm$    0.34)   &    9.46      &    8.38      &   {\bfseries  1.13}      &   31.05      &   28.14      &   {\bfseries  1.10} \\
\hline
\end{tabular}
\end{minipage}
\end{center}
\end{table*}

In Tables~\ref{tab2}-\ref{tab9},
 each testing accuracy is
  the mean and standard deviation of 100 simulations,
  while  each train time is the mean of  $100$ simulations.
The simulations show that both the original and proposed implementations always achieve the same testing accuracy,
which is then listed only once in the tables.
The speedups in Tables~\ref{tab2}-\ref{tab9} are
 ${T_{\text{original}}}/{T_{\text{proposed}}}$, i.e.,
the  ratios between the training time of the BLS using the original  implementation
 and that of the BLS using the proposed efficient implementation.
It can be seen from
Tables~\ref{tab2}-\ref{tab9}
that the speedups in total training time  of
the BLS using the proposed efficient implementation
 over  the BLS using the original  implementation
 are 1.35-1.57 on the MNIST dataset and 1.10-1.35 on the NORB dataset when $q>k$,
 while the speedups are 1.09-1.13 on the MNIST dataset and 1.07-1.09 on the NORB dataset when $q<k$.

 \begin{table*}[!t]
 \begin{center}
\begin{minipage}{\textwidth}
 \scriptsize
\renewcommand{\arraystretch}{1.3}
\setlength\tabcolsep{2pt}%
\newcommand{\tabincell}[2]{\begin{tabular}{@{}#1@{}}#2\end{tabular}}
\caption{Testing Accuracies and Training Time of the  Implementations for
BLS on NORB Dataset with $k=3100$ nodes (i.e., $2100$ enhancement and $1000$ feature nodes)
and $q=3000(<k)$ added training samples in each update} \label{tab8} \centering
\begin{tabular}{|c||c|c  c| c |c c|c|}
\hline
\multirow{4}*{ {\bfseries   \tabincell{c}{Number \\ of \\  Input \\ Patterns}}}     &  \multirow{4}*{ {\bfseries   \tabincell{c}{Testing \\ Accuracy \\ ($\% $)}}}      &\multicolumn{3}{c|}{{\bfseries  Each Additional }}       &\multicolumn{3}{c|}{{\bfseries  Accumulative   }}  \\
   &       &\multicolumn{3}{c|}{{\bfseries  Training    }}
       &\multicolumn{3}{c|}{{\bfseries  Training  }}  \\
   &       &\multicolumn{3}{c|}{{\bfseries   Time   }}
       &\multicolumn{3}{c|}{{\bfseries  Time  }}  \\
   \cdashline{3-8}
      &  &Orig.        &Impr.       & {{\bfseries   Speedups}}   &Orig.        &Impr.      &{{\bfseries Speedups}}      \\
\hline
  \bfseries 12300    &89.00     ($\pm$    0.35)   &    2.10      &    2.09      &          &    2.10      &    2.09      &       \\
\hdashline
 \bfseries  $\rightarrow$ 15300   &89.22     ($\pm$    0.37)   &    3.95      &    3.64      &   {\bfseries  1.09}      &    6.05      &    5.73      &   {\bfseries  1.06} \\
\hdashline
\bfseries  $\rightarrow$ 18300  &89.30     ($\pm$    0.36)   &    4.82      &    4.48      &   {\bfseries  1.08}      &   10.87      &   10.21      &   {\bfseries  1.06}\\
  \hdashline
 \bfseries  $\rightarrow$ 21300     &89.37     ($\pm$    0.35)   &    5.75      &    5.19      &   {\bfseries  1.11}      &   16.63      &   15.40      &   {\bfseries  1.08} \\
  \hdashline
 \bfseries  $\rightarrow$ 24300   &89.37     ($\pm$    0.35)   &    6.71      &    6.04      &   {\bfseries  1.11}      &   23.33      &   21.44      &   {\bfseries  1.09}\\
\hline
\end{tabular}
\end{minipage}
\end{center}
\end{table*}

\begin{table*}[!t]
 \begin{center}
\begin{minipage}{\textwidth}
 \scriptsize
\renewcommand{\arraystretch}{1.3}
\setlength\tabcolsep{2pt}%
\newcommand{\tabincell}[2]{\begin{tabular}{@{}#1@{}}#2\end{tabular}}
\caption{Testing Accuracies and Training Time of the  Implementations for
BLS on NORB Dataset with $k=3100$ nodes (i.e., $2100$ enhancement and $1000$ feature nodes)
and $q=2000(<k)$ added training samples in each update} \label{tab9} \centering
\begin{tabular}{|c||c|c  c| c |c c|c|}
\hline
\multirow{4}*{ {\bfseries   \tabincell{c}{Number \\ of \\  Input \\ Patterns}}}     &  \multirow{4}*{ {\bfseries   \tabincell{c}{Testing \\ Accuracy \\ ($\% $)}}}      &\multicolumn{3}{c|}{{\bfseries  Each Additional }}       &\multicolumn{3}{c|}{{\bfseries  Accumulative   }}  \\
   &       &\multicolumn{3}{c|}{{\bfseries  Training    }}
       &\multicolumn{3}{c|}{{\bfseries  Training  }}  \\
   &       &\multicolumn{3}{c|}{{\bfseries   Time   }}
       &\multicolumn{3}{c|}{{\bfseries  Time  }}  \\
   \cdashline{3-8}
      &  &Orig.        &Impr.       & {{\bfseries   Speedups}}   &Orig.        &Impr.      &{{\bfseries Speedups}}      \\
\hline
  \bfseries 16300    &89.25    ($\pm$     0.39  )  &     2.87   &    2.86   &        &    2.87   &    2.86   &          \\
\hdashline
 \bfseries  $\rightarrow$ 18300   &89.35    ($\pm$     0.39  )  &     3.65   &    3.39   &   {\bfseries  1.08}   &    6.52   &    6.25   &   {\bfseries  1.04}     \\
\hdashline
  \bfseries  $\rightarrow$ 20300  & 89.36    ($\pm$     0.38  )  &     3.96   &    3.66   &   {\bfseries  1.08}   &   10.48   &    9.91   &   {\bfseries  1.06}     \\
  \hdashline
 \bfseries  $\rightarrow$ 22300     & 89.40   ($\pm$     0.38  )  &     4.49   &    4.17   &  {\bfseries   1.08}   &   14.98   &   14.08   &   {\bfseries  1.06}     \\
  \hdashline
\bfseries  $\rightarrow$ 24300   & 89.42    ($\pm$     0.37  )  &     4.81   &    4.45   &   {\bfseries  1.08}   &   19.78   &   18.53   &   {\bfseries  1.07}     \\
\hline
\end{tabular}
\end{minipage}
\end{center}
\end{table*}


\section{Conclusion}\label{sec5}

In this paper, we speed up  the original incremental BLS on added inputs, by
 proposing an efficient implementation  of  (\ref{B_Matrix_defOriginal3241}), which is a step in the pseudoinverse computation of
 the  partitioned matrix (\ref{AxInputIncrease31232USE79547}).
  Firstly, we show that
  usually
   the condition of ${\bf{C}} ={\bf{0}}$ is satisfied in (\ref{B_Matrix_defOriginal3241}), to focus on  (\ref{B_Matrix_defOriginal3241b})
 in the original BLS implementation. To reduce the  computational complexity of (\ref{B_Matrix_defOriginal3241b}), we deduce (\ref{B_Matrix_def2bSimplebbb}) from (\ref{B_Matrix_defOriginal3241b}), and then
 utilize the inverse of a sum of matrices to deduce (\ref{B_Matrix_def2bSimpleaaa}) from (\ref{B_Matrix_def2bSimplebbb}).
 The proposed
   (\ref{B_Matrix_def2bSimplebbb}) and (\ref{B_Matrix_def2bSimpleaaa})
 are two different forms for the cases of $q \le k$ and $q > k$, respectively, where $q$  and $k$ denote  the  number of additional training samples
  and the total number of  nodes, respectively.

The proposed (\ref{B_Matrix_def2bSimpleaaa}) for $q > k$ computes only a $k \times k$ matrix inverse, instead of a $q \times q$ matrix inverse
 in the original (\ref{B_Matrix_defOriginal3241b}), and the corresponding speedup in the number of flops is $\frac{1}{2}(q/k)^3$.
 Moreover, the proposed
 (\ref{B_Matrix_def2bSimpleaaa})
 speeds up two relevant matrix multiplication operations in
 the original   (\ref{B_Matrix_defOriginal3241b}) by the speed-up  factors
 of $ql/(2 k^2)$ and $q/k$, respectively, where
  $l$  denotes the  number of training samples.
  On the other hand,
   the proposed
 (\ref{B_Matrix_def2bSimplebbb})  for  $q \le k$  speeds up one matrix multiplication operation in
 the original   (\ref{B_Matrix_defOriginal3241b}) by the speed-up  factor of $l/k$.
In
the numerical experiments,
 both  the proposed 
  and original implementations for BLS
always achieve the same testing accuracy. The speedups of the proposed efficient BLS implementation over the original BLS implementation in total training time
 are 1.35-1.57 on the MNIST dataset and 1.10-1.35 on the NORB dataset when $q>k$,
 while
  the speedups are
  1.09-1.13 on the MNIST dataset and 1.07-1.09 on the NORB dataset when $q<k$.
 We believe that the efficent implement of BLS is worthful in various applictions, for instance, the edge computing. The demand in real-time for applications has become increasingly, thus edge computing that can respond immediately exhibits its advantages.

%
%
%
%
%
%


\bibliography{sn-bibliography}


\end{document}